%% file: main.tex
\documentclass[10pt,twocolumn,letterpaper]{article}

\usepackage[pagenumbers]{wacv} 


\definecolor{wacvblue}{rgb}{0.21,0.49,0.74}
\usepackage[pagebackref,breaklinks,colorlinks,allcolors=wacvblue]{hyperref}

\usepackage{pifont} 
\newcommand{\xmark}{\ding{55}}%
\newcommand{\cmark}{\ding{51}}

\usepackage{multirow}
\usepackage{booktabs}
\usepackage{color, colortbl} 
\definecolor{light}{RGB}{179, 224, 255}
\definecolor{tabfirst}{rgb}{1, 0.7, 0.7} 
\definecolor{tabsecond}{rgb}{1, 0.85, 0.7}
\newcommand{\rc}{\cellcolor{tabsecond}}
 
\usepackage{wrapfig}
\newcommand{\ourmethod}{\text{X-Aligner}\xspace}

\def\wacvPaperID{8} 
\def\confName{EVGEN}
\def\confYear{2026}

\title{\ourmethod: Composed Visual Retrieval without the Bells and Whistles}

\author{Yuqian Zheng\\
Technical University of Munich\\
Helmholtz Munich \\
\and
Mariana-Iuliana Georgescu\\
Helmholtz Munich \\
{\tt\small georgescu\_lily@yahoo.com}
}

\begin{document}
\maketitle
\input{sec/00_abstract}
\input{sec/01_intro}
\input{sec/02_related_work}

\input{sec/03_method}
\input{sec/04_exp}
\input{sec/05_conclusions}

{
    \small
    \bibliographystyle{ieeenat_fullname}
    \bibliography{main}
}

\end{document}

%% file: sec/00_abstract.tex
\begin{abstract}
Composed Video Retrieval (CoVR) facilitates video retrieval by combining visual and textual queries. However, existing CoVR frameworks typically fuse multimodal inputs in a single stage, achieving only marginal gains over initial baseline. To address this, we propose a novel CoVR framework that leverages the representational power of Vision-Language Models (VLMs). Our framework incorporates a novel cross-attention module \textbf{\ourmethod}, composed of cross-attention layers that progressively fuse visual and textual inputs and align their multimodal representation with that of the target video. To further enhance the representation of the multimodal query, we incorporate the caption of the visual query as an additional input.  The framework is trained in two stages to preserve the pretrained VLM representation. In the first stage, only the newly introduced module is trained, while in the second stage, the textual query encoder is also fine-tuned. We implement our framework on top of BLIP-family architecture, namely BLIP and BLIP-2, and train it on the Webvid-CoVR data set. In addition to in-domain evaluation on Webvid-CoVR-Test, we perform zero-shot evaluations on the Composed Image Retrieval (CIR) data sets CIRCO and Fashion-IQ. Our framework achieves state-of-the-art performance on CoVR obtaining a Recall@1 of 63.93\% on Webvid-CoVR-Test, and demonstrates strong zero-shot generalization on CIR tasks.

\end{abstract}

%% file: sec/01_intro.tex

\section{Introduction}
\label{sec:intro} 

The rapid expansion of media content demands for more sophisticated retrieval systems capable of handling complex search queries, such as multimodal ones. Traditional text- or image-based retrieval methods lack the ability to retrieve content involving compositional changes. Therefore, Composed Video Retrieval (CoVR) and Composed Image Retrieval (CIR) have become crucial tasks, enabling users to find specific visual content by combining a reference image or video with a natural language modification query. The goal of a composed visual retrieval is to identify the visual target that best matches the visual query (image or video) modified by the corresponding text query.  Several state-of-the-art approaches~\cite{Karthik-ICLR-2024,Li-CVPR-2025} propose training-free techniques that leverage Large Language Models (LLMs) or large vision models for zero-shot composed visual retrieval. However, these methods often underperform because they lack explicit training on how to precisely transform a visual query based on a textual query. Other works~\cite{Thawakar-CVPR-2024,Ventura-AAAI-2024, thawakar2025beyond} adapt Vision-Language Models (VLMs) to composed visual retrieval by fine-tuning them on CIR or CoVR data sets.

\begin{figure}
  \centering
  \includegraphics[width=0.99\linewidth]{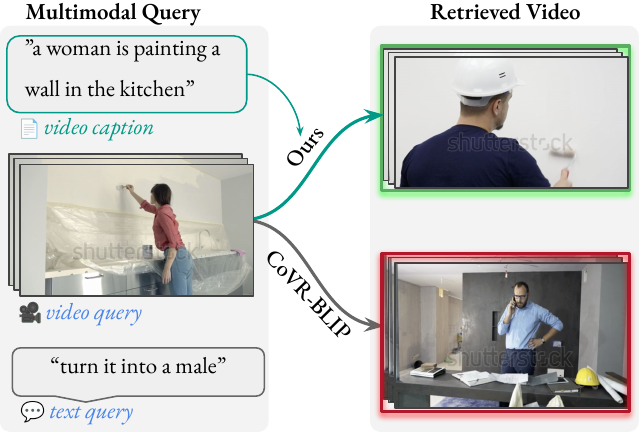}
  \caption{We propose a novel framework that leverages multi-stage cross-attention and video captions to accurately align multimodal queries.
  As shown, the single-stage baseline CoVR-BLIP~\cite{Ventura-AAAI-2024} fails interpret the ``turn it into a male'' instruction correctly, retrieving a video with a male subject. In contrast, our method fuses the visual, textual, and caption inputs to retrieve the correct video of a man painting. Our approach achieves state-of-the-art performance, such as a Recall@1 of 63.93\% on Webvid-CoVR-Test.
  }\label{fig:teaser_intro}
\end{figure}

For example, Ventura \etal~\cite{Ventura-AAAI-2024} freeze the vision encoder of BLIP~\cite{Li-ICML-2022} and fine-tune its multimodal encoder. However, performance gains on the CoVR benchmarks have remained limited, with subsequent state-of-the-art models only marginally outperforming the baselines proposed by Ventura~\etal~\cite{Ventura-AAAI-2024}. We observe that information from the input query is processed only once by the model encoders, without being integrated across multiple stages. Single-stage input integration limits the models' ability to iteratively refine and align multimodal representations, potentially leading to suboptimal understanding of complex queries, where subtle textual modifications require deeper interaction with the visual features.

In this work, we address the Composed Video Retrieval task by leveraging the rich knowledge encoded in VLMs.  We propose \textbf{\ourmethod}, a module that combines visual and textual inputs through multiple cross-attention layers to improve the alignment between their representations. To further enhance the representation, the input embeddings are integrated across multiple stages. The caption of the visual query, automatically generated by InternVL-G~\cite{Chen-CVPR-2024}, is also incorporated to enrich semantic understanding by providing high-level visual context in language space that complements the visual features.
To better adapt VLMs to the CoVR task, we design a novel two-stage training framework. In the first stage, we exclusively train the newly added cross-attention components (\ourmethod). In the second stage, we jointly fine-tune the textual query encoder along with \ourmethod, while keeping the remaining model parameters frozen to prevent catastrophic forgetting of the pretrained VLM's original multimodal knowledge.

The proposed framework is built upon two widely used VLMs, namely BLIP~\cite{Li-ICML-2022} and BLIP-2~\cite{Li-ICML-2023}. Its effectiveness is validated through extensive experiments on the Webvid-CoVR data set~\cite{Ventura-AAAI-2024}, where it achieves a state-of-the-art Recall@1 of 63.93\% on the test set. As illustrated in Figure~\ref{fig:teaser_intro}, our framework successfully combines multimodal inputs to retrieve the target video, showcasing its ability to handle complex compositional queries. Furthermore, the strong generalization of the model is demonstrated through zero-shot CIR evaluations on the FashionIQ~\cite{Wu-CVPR-2021} and CIRCO~\cite{Baldrati-ICCV-2023} data sets.




We summarize our contributions as follows. 

\begin{itemize}
    \item We introduce a simple, yet powerful framework that aligns visual and textual inputs through progressive cross-modal reasoning, effectively adapting pretrained VLMs for Composed Video Retrieval. 
    \item We achieve state-of-the-art performance on the WebVid-CoVR dataset, surpassing existing baselines by a significant margin.
    \item We demonstrate that the representations learned by our framework generalize well to Composed Image Retrieval, achieving competitive performance in zero-shot CIR task.

\end{itemize}

%% file: sec/02_related_work.tex
\section{Related Work}

\noindent
\textbf{Composed Image Retrieval} (CIR): 
There has been a growing interest in CIR~\cite{Vo-CVPR-2019} in the research community in recent years.
Several CIR methods~\cite{Baldrati-CVPR-2022, Saito-CVPR-2023,Baldrati-CVPR-2023, Gu-TMLR-2024,Karthik-ICLR-2024,Gu-CVPR-2024,Zhang-PMLR-2024,Zhou-Arxiv-2024} leverage CLIP~\cite{Radford-ICML-2021} to encode the query (reference) image and query text into a shared embedding space for target image retrieval. Specifically, Gu~\etal~\cite{Gu-CVPR-2024} introduced a language-only training framework for CIR that maps images to text representations, thereby reformulating CIR as a text-to-image retrieval task. 

Several other works have proposed training-free methods~\cite{Karthik-ICLR-2024,Yang-ICM-2024,Tang-arxiv-2024,Li-CVPR-2025,Luo-WC-2025}. For instance, Karthik~\etal~\cite{Karthik-ICLR-2024} proposed CIReVL, a framework that utilizes LLMs to generate a target image caption by combining the text query and the visual input caption, thus performing text-to-image retrieval. On the other hand, Li~\etal~\cite{Li-CVPR-2025} approached the CIR task from an image-to-image retrieval perspective, employing an image generation model to synthesize proxy images based on multimodal queries. To further improve CIR performance, Levy~\etal~\cite{Levy-AAAI-2024} introduced the Large-Scale Composed Image Retrieval (LaSCo) dataset and proposed a new CIR model based on BLIP~\cite{Li-ICML-2022}. While the aforementioned methods are specifically designed for CIR, however, our work presents a framework that is fine-tuned on a video data set and demonstrates its generalization capabilities through performing Zero-Shot (ZS) Composed Image Retrieval.

\noindent
\textbf{Composed Video Retrieval} (CoVR): The Composed Video Retrieval task was introduced by Ventura~\etal~\cite{Ventura-AAAI-2024}, who presented the first CoVR framework by adapting BLIP~\cite{Li-ICML-2022} and training it on WebVid-CoVR data set, integrating multimodal inputs via summation or cross-attention. Subsequently, CoVR-2~\cite{Ventura-TPAMI-2024} extended this approach using BLIP-2~\cite{Li-ICML-2023}, achieving improved alignment and retrieval performance. Furthermore, Hummel~\etal~\cite{Hummel-ECCV-2024} introduced EgoCVR, a temporal reasoning benchmark for the egocentric CoVR task. In addition, Wu~\etal~\cite{Wu-ICLR-2025} proposed FDCA, a framework focused on feature-level disentanglement trained on their FineCVR-1M data set.

Similar to our approach, Thawakar~\etal~\cite{Thawakar-CVPR-2024} built their framework based on CoVR~\cite{Ventura-AAAI-2024}, along with an additional detailed caption of the visual query. Inspired by this, we also employ captioning models to obtain visual query descriptions. However, we employ concise captions instead of lengthy and detailed ones. Unlike prior methods, we develop a lightweight yet effective module that jointly encodes the three inputs across multiple stages, achieving state-of-the-art performance on the WebVid-CoVR-test benchmark. More recently, Thawakar~\etal~\cite{thawakar2025beyond} introduced the Dense-WebVid-CoVR dataset, which focuses on fine-grained modifications through dense textual descriptions. Although Thawakar~\etal~\cite{thawakar2025beyond}'s approach emphasizes high-density information, our framework demonstrates that progressive fusion with concise captions can achieve superior alignment without the computational overhead of processing dense text.

%% file: sec/03_method.tex

\begin{figure*}
\centering
  \includegraphics[width=0.99\linewidth]{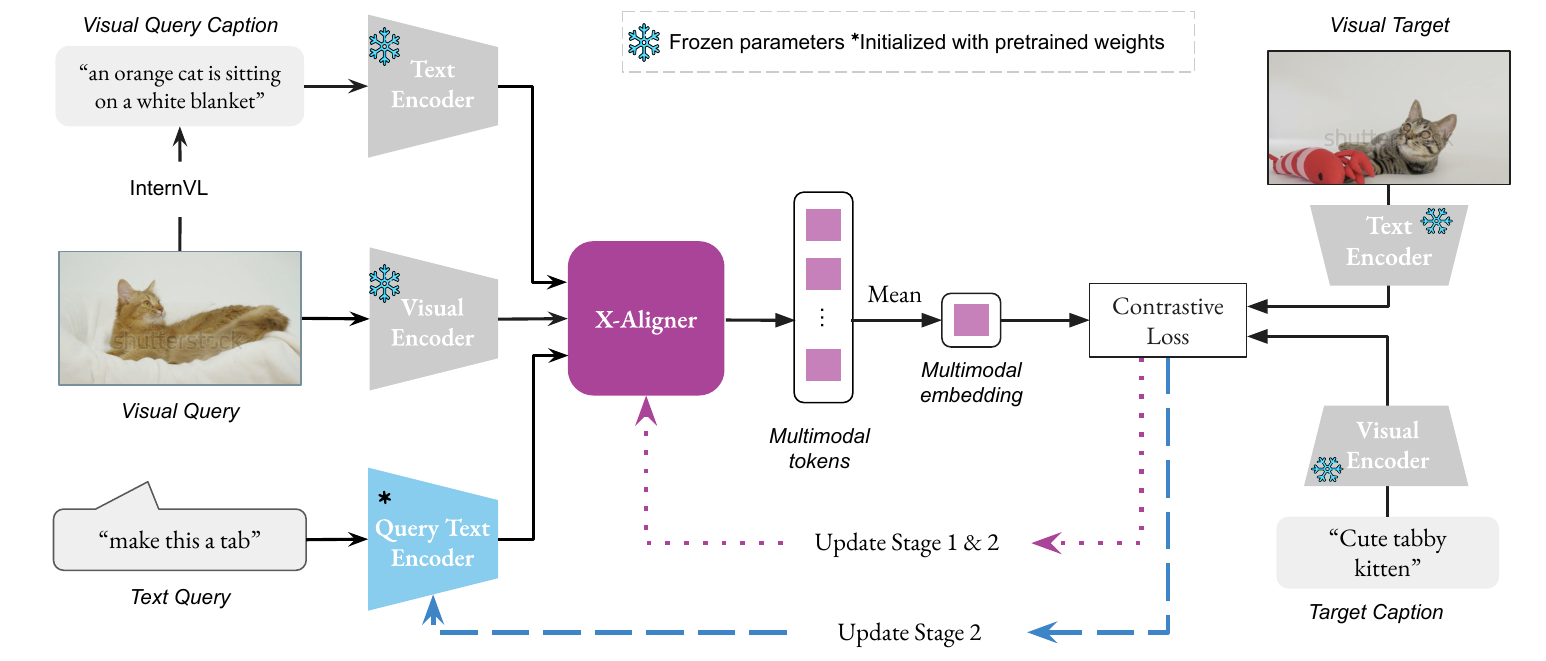}
\caption{We present an overview of our framework. Our fusion adapter \textbf{\ourmethod} is integrated on top of the embeddings extracted from the Vision-Language Models. The \textit{Text Encoder} and \textit{Query Text Encoder} share the same parameters in Stage 1. The resulting multimodal embedding is then aligned with the target embeddings (visual and textual) using contrastive loss. Components updated during each training stage are indicated with dashed and dotted lines. ``Tab'' is the nickname used for tabby cat.} 
\label{fig:teaser}
\end{figure*}

\section{Method}

\subsection{Problem Definition}
Composed Video Retrieval (CoVR) involves retrieving a target video given a multimodal query consisting of a visual reference input and a text modification. The ground truth video is defined as the one that best matches the visual input conditioned on the text modification. Formally, given a gallery of videos $V$, a text modification $q_t$, and visual reference input $q_v$, our objective is to learn a mapping function that integrates these inputs, along with an auxiliary visual caption, to retrieve the target video $v \in V$ that most accurately reflects the desired  modification within temporally dynamic scenes, based on both the original visual input and the accompanying text description.  
To perform the retrieval, it requires a text encoder $f_{t}$ to generate the textual embeddings, a visual encoder $f_{v}$ to produce visual embeddings, and optionally a multimodal fusion module $f_{tv}$ to integrate these two modalities. The target video $v$ is identified as the one that maximizes the similarity between the target representation $f_v(v)$ and the fused query representation $f_{tv}(f_t(q_t), f_v(q_v))$. In practice, cosine similarity is commonly adopted as the similarity metric.

\subsection{Baseline Framework}
\begin{figure*}
    \centering
  \includegraphics[width=0.9\linewidth]{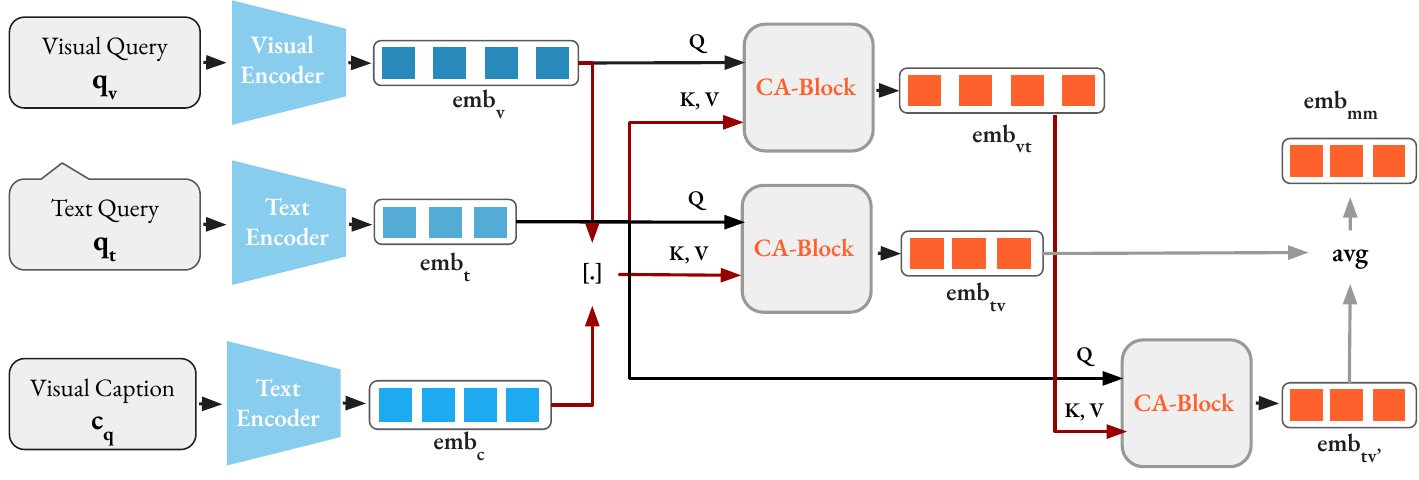}
\caption{We present the \textbf{components of \ourmethod}. We obtain the multimodal embedding $emb_{tv}$ by enriching the text embedding with information from the visual query ($emb_v$ and $emb_c$). We progressively integrate multimodal input by applying cross-attention between the text embedding $emb_t$ and the multimodal embedding $emb_{vt}$. The final embedding  $emb_{mm}$ is computed as the average of the embeddings produced by each components, namely $emb_{tv}$ and $emb_{tv'}$. We depict with black arrows the query, and with red arrows the keys and values.}
\label{fig:components}
\end{figure*}

Building upon the insights from Ventura~\etal~\cite{Ventura-TPAMI-2024}, our framework explores adapting VLMs to the CoVR setting. A typical VLM comprises of two encoders, a vision encoder $f_v$ and a text encoder $f_t$, which are jointly optimized during pretraining to produce aligned representations. To tailor the model for the CoVR task, we design the framework to  take as input a visual query $q_v$, a text query $q_t$, and an auxiliary visual caption $q_c$. These inputs are encoded and subsequently fused into a joint representation $f_{tv}(f_t(q_t), f_v(q_v))$, as shown in Figure~\ref{fig:components}.  

\subsection{Our Framework}
Aiming to fully exploit the potential of multimodal queries, we propose a multi-stage fusion strategy (\ourmethod) paired with a novel two-stage fine-tuning approach. An overview of the proposed framework is illustrated in Figure~\ref{fig:teaser}.

As described above, our framework leverages a VLM architecture comprising a text encoder $enc_t$ and a vision encoder $enc_v$. Specifically, the embedding of the text query $q_t$ is computed as $emb_{t} = enc_t(q_t)$, while the representation of the visual input is extracted as $emb_{v} = enc_v(q_v)$. We denote a transformer block incorporating cross-attention as $\texttt{\textbf{CA-Block}(q, kv)}$, where $q$ denotes the query tokens and $kv$ represents the key and value tokens attented to in the cross-attention layer. Each transformer block follows a standard architecture where a cross-attention (CA) layer is inserted between the self-attention (SA) layer and the feed-forward network (FFN). The fusion of multimodal information is subsequently performed through two distinct paths.

\textbf{Leveraging Captions for Multimodal Fusion Enhancement.} 
Firstly, we generate the caption $q_{c}$ of the visual query using InternVL~\cite{Chen-CVPR-2024}, a large-scale vision-language model capable of visual captioning, and encode it as $emb_{c} = enc_t(q_{c})$. To incorporate this supplementary semantic information, we employ two stacked transformer blocks that enrich the original text query representation $emb_{t}$ by attending to both the caption tokens $emb_{c}$ and the visual query tokens $emb_{v}$. The final output of this two-block stack serves as the multimodal query representation $emb_{tv}$, computed as $emb_{tv} = \texttt{\textbf{CA-Block}}(q=emb_{t}, kv= \big[emb_{c}, emb_{v}\big])$, where $\big[\, \cdot \,\big]$ denotes the concatenation operation along the sequence dimension. This component is illustrated in  Figure~\ref{fig:components}.



\textbf{Bi-directional Cross-Attention for Query Refinement.}
As a core part of the query refinement process, we introduce a dual cross-attention mechanism to facilitate deeper interaction between the visual and textual modalities.
Initially, the visual query representation is  refined by incorporating information from the textual query. This is achieved through a cross-attention operation within the $\texttt{\textbf{CA-Block}}$, which produces the a text-conditioned visual representation: $emb_{vt}=\texttt{\textbf{CA-Block}}(q=emb_{v}, kv=emb_{t})$.
Following this, we perform the second multimodal fusion step by integrating the text query embedding $emb_{t}$  with the newly formed multimodal embedding $emb_{vt}$. 
To capture complementary information from \( \text{emb}_{vt} \), an additional transformer block (architecturally identical to the previous one) is employed using \( \text{emb}_{t} \) as the query and \( \text{emb}_{vt} \) as the key and value. This results in a refined multimodal embedding:
${emb}_{tv'} = \texttt{\textbf{CA-Block}}(q = emb_{t}, kv=emb_{vt})$. This bi-directional interaction enables further refinement of the query text representation, building upon the multimodal enrichment from the previous stage. A visual representation of this component is provided in Figure~\ref{fig:components}.

The final multimodal query representation  is obtained by averaging (because we only select the class token as the input representation) the outputs from the two fusion paths, resulting in the joint embedding:
\[
\text{emb}_{mm} = \frac{\text{emb}_{tv} + \text{emb}_{tv'}}{2}.
\]
In our framework, the first path, the embedding $\text{emb}_{tv}$ provides high-level semantic context via captions, while the second path ensures  $\text{emb}_{tv'}$ fine-grained visual-textual alignment through direct bi-directional interaction between visual and language representation. Together, these two components enable the model to leverage both caption-guided semantic enrichment ($\text{emb}_{tv}$) and bidirectional cross-modal interactions ($\text{emb}_{tv'}$), facilitating a more comprehensive and context-aware multimodal representation for retrieval. 

\textbf{Two-stage Fine-tuning: From \ourmethod Training to Joint Optimization with Text Query.} Our framework is fine-tuned in two stages, as illustrated in Figure~\ref{fig:teaser}. During the first stage, exclusively the parameters of \ourmethod are updated, while pretrained the visual and textual encoders remain frozen. This design allows the newly introduced components to adapt effectively to the embedding distributions of the pretrained VLM. This also prevents the degradation of pretrained VLM, eliminating catastrophic forgetting.

In the second stage, we unfreeze the query text  encoder to further refine the text query representation $emb_{t}$ while the caption representation, extracted using the same pretrained encoder, remains frozen. This joint optimization allows the text encoder for the modification query to capture task-specific semantics in CoVR, while the frozen caption input preserves the general semantic knowledge learned during pretraining.

\begin{table*}[t]
  \centering
  \caption{We report performance on WebVid-CoVR-Test~\cite{Ventura-AAAI-2024} using Recall@1 (R@1), Recall@5 (R@5) and Recall@10 (R@10) as evaluation metrics. All compared methods are built upon Vision-Language Models and fuse the multimodal input either by averaging (avg) the embeddings, applying cross-attention (CA), or by using our proposed \ourmethod method. All models are finetuned on the WebVid-CoVR-Train data set. The highest-performing result is highlighted in bold. Our framework obtains the top performance regardless of the~metric.}
  \label{tab:covr}
  \begin{tabular}{cllcccc}
    \toprule
    \bf Training & \bf Framework & \bf Backbone & \bf Fusion & \bf R@1 & \bf R@5 & \bf R@10 \\
    \midrule
    \multirow{3}{*}{Not finetuned}
    & -- & CLIP & Avg & 44.37 & 69.13 & 77.62 \\
    & -- & BLIP & Avg & 45.46 & 70.46 & 79.54 \\
    & -- & BLIP-2 & Avg & 45.66 & 71.71 & 81.30 \\
    \midrule
    \multirow{5}{*}{Finetuned}
    & CoVR~\cite{Ventura-TPAMI-2024} & BLIP   & CA & 55.95 & 81.22 & 89.05 \\
    & CoVR-2~\cite{Ventura-TPAMI-2024} & BLIP-2 & CA & 59.82 & 83.84 & 91.28 \\
    & ECDE~\cite{Thawakar-CVPR-2024}      & BLIP   & CA & 60.12 & 84.32 & 91.27 \\
    & WebVid-CoVR~\cite{thawakar2025beyond}      & BLIP-2   & CA & 60.40 & 84.50 & 91.40 \\
    & Dense-CoVR~\cite{thawakar2025beyond}      & BLIP-2   & CA & 63.80   & 87.50 & 92.40 \\
    & \multirow{2}{*}{  \bf Ours (Stage 1)} 
    &  BLIP   &   \ourmethod & 61.89 &  84.55 &  90.92 \\
    &     &   BLIP-2 &  \ourmethod  &   62.79 &   86.54 &   92.06 \\
    & \multirow{2}{*}{  \bf Ours} 
    & \rc BLIP   &  \rc \ourmethod & \rc 63.50 &  \rc 85.95 & \rc 91.59 \\
    &     & \rc  BLIP-2 & \rc  \ourmethod  & \rc  \textbf{63.93} & \rc  \textbf{87.01} & \rc  \textbf{92.41} \\
    \bottomrule
  \end{tabular}
\end{table*}

In both stages, the model parameters are optimized using the HN-NCE~\cite{rdk+23} loss. Consistent with to CoVR-2~\cite{Ventura-TPAMI-2024}, this loss term aligns the joint representation with both the embedding of the target video $y_v$ and the embedding of its corresponding ground-truth caption $y_c$, which is provided by the Webvid-CoVR dataset as supervision. The ground-truth caption $y_c$ is employed only during training. Both loss components are weighted equally during training. 


%% file: sec/04_exp.tex
 
\begin{table*}
  \caption{Zero-Shot Performance on FashionIQ~\cite{Guo-CVPR-2021}. Our models are finetuned on the WebVid-CoVR-Train data set~\cite{Ventura-AAAI-2024}. We compare our framework to training-free methods and to those trained on a comparable number of sample. The highest-performing result is highlighted in bold, while the second-best is underlined. Our framework shows strong zero-shot generalization capabilities.}
  \label{tab:fiq}
  \centering
  \begin{tabular}{llcc|cc|cc|cc}
    \hline
     \bf Method &    \bf Backbone &  \multicolumn{2}{c}{\bf Dress}    &     \multicolumn{2}{c}{\bf Shirt}   &   \multicolumn{2}{c}{\bf Toptee}  &   \multicolumn{2}{c}{\bf Average}\\
     \cline{3-10}
                &                 &   R@10 & R@50  &   R@10 & R@50  &   R@10 & R@50  &   R@10 & R@50 \\
     
     \hline
 
     Random    &        --         &  0.26  &  1.31 &  0.16  &  0.79 &  0.19  &  0.95 & 0.06  &  0.32  \\
     CompoDiff~\cite{Gu-TMLR-2024} &  ViT-L  & 32.24  & 46.27 & 37.69  & 49.08 & 38.12 & 50.57 &  36.02 & 48.64 \\
     Pic2Word~\cite{Saito-CVPR-2023} &  CLIP (ViT-L) & 20.00  & 40.20 & 26.20  &  43.60 & 27.90 & 47.40 & 24.70 & 43.70 \\
     SEARLE-XL~\cite{Baldrati-CVPR-2023} &  CLIP (ViT-L) & 26.89  & 45.58 & 20.48  & 43.13  & 29.32 & 49.97 & 25.56 & 46.23 \\
     CIReVL~\cite{Karthik-ICLR-2024} & CLIP (ViT-L) & 24.79 & 44.76 & 29.49 & 47.40 & 31.36 & 53.65 & 28.55 & 48.57 \\
     LinCIR~\cite{Gu-CVPR-2024} & ViT-L  & 20.92  & 42.44 & 29.10  & 46.81 & 28.81  & 50.18 & 26.28 & 46.49 \\
     CoVR-BLIP~\cite{Ventura-AAAI-2024} & BLIP (ViT-L)              & 21.95  & 39.05 & 30.37  & 46.12 & 30.78  & 48.73 & 27.70 & 44.63 \\
     CoVR-BLIP-2~\cite{Ventura-TPAMI-2024} & BLIP-2  (ViT-L)         & --  & -- & --  & -- & --  & -- &  \bf 36.81 &  \bf 56.70 \\
     ECDE~\cite{Thawakar-CVPR-2024} & BLIP (ViT-L) & 24.57  & 40.93 & 33.12  & 48.42 & 33.16  & 50.24 & 30.28 &  46.53 \\
     WebVid-CoVR~\cite{thawakar2025beyond} & BLIP (ViT-L) & 21.08  & 38.26 &  22.18  &  36.72 &  25.06 &  44.28 & 22.77 & 39.75 \\
     Dense-CoVR~\cite{thawakar2025beyond} & BLIP (ViT-L) & 26.12 & 42.88 & 35.32 & 49.92 & 35.44 & 51.66 & 32.29 & 48.15
     \\
     \multirow{2}{*}{\bf Ours} 
     & \rc BLIP (ViT-L) & \rc 30.15 & \rc 50.15 & \rc 37.40 & \rc 55.34 & \rc 39.00 & \rc 60.49 &  \rc 35.52 & \rc 55.33 \\
                               & \rc BLIP-2 (ViT-L) &  \rc 32.38 & \rc 54.11 &  \rc 39.61 & \rc 57.60 & \rc 37.12 & \rc 58.20 & \rc  \underline{36.37} & \rc  \underline{56.63} \\

     \hline
  \end{tabular}
\end{table*}

\section{Experiments}

\noindent
\textbf{Data Sets.}
\textbf{WebVid-CoVR} is the first data set designed for the Composed Video Retrieval task, introduced by Ventura \etal~\cite{Ventura-AAAI-2024}. It comprises 1.6 million automatically generated training triplets, each consisting of a text query, a video query, and a corresponding target video. In addition, the WebVid-CoVR-Test set is human-annotated and contains 2,556 high-quality triplets. Following previous work~\cite{Ventura-AAAI-2024,Thawakar-CVPR-2024,Ventura-TPAMI-2024}, we adopt Recall@1 (R@1), Recall@5 (R@5), and Recall@10 (R@10) as evaluation~metrics.

The generalization ability of our method is validated through zero-shot Composed Image Retrieval on FashionIQ~\cite{Wu-CVPR-2021}  validation split and CIRCO~\cite{Baldrati-ICCV-2023} test split.

\noindent
\textbf{FashionIQ}~\cite{Wu-CVPR-2021}  is a benchmark data set for composed image retrieval, containing images of fashion products categorized into three classes: Shirts, Dresses, and Tops/Tees. For each query, a pair of query and target images is constructed based on title similarity, with corresponding text modifications crafted to describe the visual difference. The validation split comprises 6,016 queries and a gallery of 15,415 images. To achieve a fair comparison with previous methods~\cite{Ventura-AAAI-2024,Ventura-TPAMI-2024,Thawakar-CVPR-2024}, we report retrieval performance using Recall@10 and Recall@50 on the validation split, evaluating both per-category and average results.

\noindent
\textbf{CIRCO}~\cite{Baldrati-ICCV-2023} is a large-scale, open-domain dataset for composed image retrieval, constructed from real-world images in the COCO 2017 unlabeled set~\cite{Lin-ECCV-2014}. Each CIRCO query is constructed based on a pair of visually similar images, accompanied by a human-authored relative caption and supplementary annotations that highlight shared attributes to reduce ambiguity. The dataset contains 1,020 queries, split into 220 for validation and 800 for testing, and uses the full COCO image set (120K images) as the retrieval gallery, offering a rich set of visually similar distractors that increase the difficulty and discriminative requirements of the retrieval task. To account for the multiple ground-truth targets per query in CIRCO, mean Average Precision (mAP) is used for evaluation. Results are reported at mAP@5, mAP@10, mAP@25, mAP@50.


 
\textbf{Implementation Details.}
The new video captions are generated using InternVL-G~\cite{Chen-CVPR-2024}, a powerful vision-language foundation model. Our designed \ourmethod incorporates one or two randomly initialized transformer layers based on the BERT architecture. Inspired by prior work~\cite{Ventura-AAAI-2024, Thawakar-CVPR-2024,Ventura-TPAMI-2024}, we adopt pre-trained text and visual encoders from the BLIP family, namely BLIP~\cite{Li-ICML-2022} and BLIP-2~\cite{Li-ICML-2023}, both of which were fine-tuned on the COCO dataset~\cite{Lin-ECCV-2014} for text-image retrieval tasks. 

The model is trained exclusively on the WebVid-CoVR training set. For caption encoding, both BLIP- and BLIP-2-based backbones employ the pre-trained text encoder to process the input caption text independently, without incorporating any visual features. In the BLIP-based variant, the query text is encoded similarly using a standalone text encoder. In contrast, the BLIP-2-based variant adopts the CoVR-2~\cite{Ventura-TPAMI-2024} architecture, where the text query encoder receives both the textual input and learnable query tokens, and includes cross-attention layers that integrate visual embeddings. During the second stage of training, these cross-attention layers are kept frozen to ensure that only the text-related parameters are updated.

Training is performed in two stages, each for 10 epochs. The learning rate is set to 2e-4 in the first stage and reduced to 1e-05 in the second. For BLIP-based models, we use a total batch size of 2048 (512 per GPU), while for BLIP-2-based models, the batch size is set to 1024 (256 per GPU). Since both VLMs are designed to process a single image at a time, both video inputs and targets are handled by encoding each frame independently and averaging their embeddings to obtain a unified video representation.
All experiments are conducted using 4 NVIDIA H100 GPUs.

\subsection{Composed Video Retrieval Results}

\textbf{Comparison to State-Of-The-Art CoVR.} Results on the WebVid-CoVR data set in terms of Recall@1 (R@1), Recall@5 (R@5), and Recall@10 (R@10) are presented in Table~\ref{tab:covr}. The comparison includes training-free baselines proposed by Ventura~\etal~\cite{Ventura-TPAMI-2024}, where retrieval is performed by averaging the embeddings of the middle video frame and the text query. These baselines exhibit relatively low performance due to their simple fusion technique (average or cross-attention), with the BLIP-2~\cite{Li-ICML-2023}-based variant achieving an R@1 of only 45.66. The previous state-of-the-art on this benchmark was established by ECDE~\cite{Thawakar-CVPR-2024}, which achieved an R@1 of 60.12.

Our proposed framework, equipped with the \ourmethod module for multimodal input fusion, outperforms all prior baselines by a substantial margin. Notably, even in the Stage 1 setting, where the parameters responsible for encoding the modification text query are kept fixed, our model already outperforms all existing methods. Furthermore, the improvement observed in Stage 2 validates the importance of adapting the text-query encoder to task-specific, while keeping the caption encoder frozen to preserve the common (out-of-domain) knowledge. When using the BLIP backbone, our model achieves R@1 of 63.50, R@5 of 85.95, and R@10 of 91.59, surpassing CoVR-2~\cite{Ventura-TPAMI-2024}, ECDE~\cite{Thawakar-CVPR-2024} and Dense-CoVR~\cite{thawakar2025beyond} despite its architectural simplicity. Switching to the BLIP-2 backbone yields further improvements, with an R@1 of 63.93, R@5 of 87.01, and R@10 of 92.41, outperforming all previous methods on the WebVid-CoVR test set. 

The results demonstrate that our framework is robust to backbone selection, consistently outperforming more complex architectures~\cite{Thawakar-CVPR-2024,thawakar2025beyond}.


\subsection{Zero-Shot Composed Image Retrieval Results}

\begin{table*}
    
  \caption{Zero-Shot performance on CIRCO~\cite{Baldrati-ICCV-2023}.  Our models are finetuned on the WebVid-CoVR-Train data set~\cite{Ventura-AAAI-2024}. We compare our framework with training-free methods or methods trained on comparable number of samples. The highest-performing result is highlighted in bold, while the second-best is underlined. Our framework shows strong zero-shot generalization capabilities.}
  \label{tab:circo}
  \centering
    \resizebox{0.8\linewidth}{!}{
  \begin{tabular}{llcccc}
    \hline
     \bf Method &    \bf Backbone            &  \bf mAP@5  & \bf mAP@10 & \bf mAP@25 & \bf mAP@50 \\
      \hline %
      Pic2Word~\cite{Saito-CVPR-2023}    &    CLIP (ViT-L)                &  8.70        & 9.50        & 10.60       & 11.30 \\
      SEARLE~\cite{Baldrati-CVPR-2023}      &    CLIP (ViT-L)                 & 11.70       & 12.70       & 14.30       & 15.10 \\
      CompoDiff~\cite{Gu-TMLR-2024}   &     CLIP (ViT-L)                 & 12.60        & 13.40       & 15.80       & 16.40 \\
      CIReVL~\cite{Karthik-ICLR-2024}      &     CLIP (ViT-L)                 & 18.57       & 19.01      & 20.89      &  21.80\\ 
      LinCIR~\cite{Gu-CVPR-2024}      &    CLIP (ViT-L)               & 12.60        & 13.60       & 15.00       & 15.90 \\
      CoVR-BLIP~\cite{Ventura-AAAI-2024} &     BLIP (ViT-L)                & 21.43      & 22.33      & 24.47      & 25.48 \\
      CoVR-BLIP-2~\cite{Ventura-TPAMI-2024}&     BLIP-2 (ViT-L)               &  \bf 28.88       & --      & --      & -- \\
      \multirow{2}{*}{\bf Ours}   &   \rc  BLIP (ViT-L)                 & \rc \underline{25.72}   &  \rc \underline{26.86}          & \rc 29.16    & \rc \underline{30.20}    \\
                                  &   \rc  BLIP-2 (ViT-L)               &  \rc   25.11   &  \rc 26.49          & \rc  \underline{28.83}    &  \rc 29.98    \\

    \hline
  \end{tabular}
  }
\end{table*}

\begin{table*}[t]
  \centering
  \caption{Ablation results evaluating the contribution of incorporating the input query caption into \ourmethod. We report Composed Video Retrieval results on WebVid-CoVR-Test~\cite{Ventura-AAAI-2024} (CoVR), and Zero-Shot Composed Image Retrieval on FashionIQ~\cite{Wu-CVPR-2021} and CIRCO~\cite{Baldrati-ICCV-2023}. Adding the input query caption generally improves the performance.} 
  \label{tab:ablation_adding_caption}
  \begin{tabular}{ccccc}
    \toprule
     \bf Backbone & \bf Caption  & \bf CoVR R@1 & \bf FashionIQ R@10 & \bf CIRCO mAP@5 \\
    \midrule 
    \multirow{2}{*}{BLIP}  &  \xmark  & \rc 63.89 & 35.25 & 25.61 \\
                           &    \cmark & 63.50 & \rc 35.52 & \rc 25.72 \\
    \midrule 
    \multirow{2}{*}{BLIP-2} & \xmark   & 62.99 &  \rc  36.41 & 24.57 \\
                          & \cmark    &  \rc 63.93 & 36.37 & \rc  25.11 \\
    \bottomrule
  \end{tabular}
\end{table*}

To evaluate the generalization capability of  \ourmethod beyond the video domain, we conduct zero-shot (ZS) composed image retrieval (CIR) experiments on two datasets, namely FashionIQ~\cite{Guo-CVPR-2021} and CIRCO~\cite{Baldrati-ICCV-2023}. We generated the embeddings using the model trained exclusively on the WebVid-CoVR training set, without further fine-tuning. This setup allows us to assess the transferability of video-learned representations to static image retrieval tasks. The results are summarized in Tables~\ref{tab:fiq} and~\ref{tab:circo}.

\noindent
\textbf{FashionIQ.} The zero-shot CIR results on the FashionIQ~\cite{Guo-CVPR-2021} dataset are presented in Table~\ref{tab:fiq}. We report the results in terms of Recall@10 (R@10) and Recall@50 (R@50). Our framework, based on the BLIP backbone, attains an average R@10 of 35.52 and R@50 of 55.33, surpassing the ECDE method~\cite{Thawakar-CVPR-2024}, which achieves an R@10 of 30.28. The average performance obtained with BLIP surpasses several methods~\cite{Ventura-AAAI-2024,Thawakar-CVPR-2024, thawakar2025beyond} that employed the same backbone model by almost 13\% in terms of R@10, validating the improvement brought by \ourmethod.

For a direct comparison, we refer to the CoVR-BLIP-2~\cite{Ventura-TPAMI-2024} result that is finetuned on the same training set. Replacing BLIP with the more advanced BLIP-2 backbone further boosts the R@10 score to 36.37, narrowing the gap with the current state-of-the-art CoVR-BLIP-2~\cite{Ventura-TPAMI-2024}, whose corresponding performance reaches 36.81. Despite its relatively simple architecture, our framework demonstrates strong zero-shot CIR performance and surpasses several ZS-CIR baselines~\cite{Gu-TMLR-2024,Saito-CVPR-2023,Baldrati-CVPR-2023,Karthik-ICLR-2024,Gu-CVPR-2024,Thawakar-CVPR-2024}.

\noindent
\textbf{CIRCO.} We report the zero-shot CIR results on the CIRCO~\cite{Baldrati-ICCV-2023} test set in Table~\ref{tab:circo}, using mean Average Precision (mAP) at various ranks (mAP@5, @10, @25, @50) as the evaluation metric. Our framework achieves the second-highest mAP@5 score of 25.72, demonstrating superior performance over several advanced methods such as SEARLE~\cite{Baldrati-CVPR-2023}, CIReVL~\cite{Karthik-ICLR-2024}, and LinCIR~\cite{Gu-CVPR-2024}. In particular, when both methods are based on the BLIP backbone, our framework improves the mAP@5 score by 4.29 points over CoVR-BLIP~\cite{Gu-CVPR-2024} (25.72 vs. 21.43), underscoring the benefit of our design under identical encoder settings. 

It is important to emphasize that while our framework was explicitly designed for CoVR and trained on video data, it still surpasses methods trained specifically on image data~\cite{Saito-CVPR-2023, Baldrati-CVPR-2023, Karthik-ICLR-2024, Gu-CVPR-2024}. This validates our the capacity of our framework for cross-domain transfer, rather than just in-domain generalization.

\begin{figure}
  \includegraphics[width=0.98\linewidth]{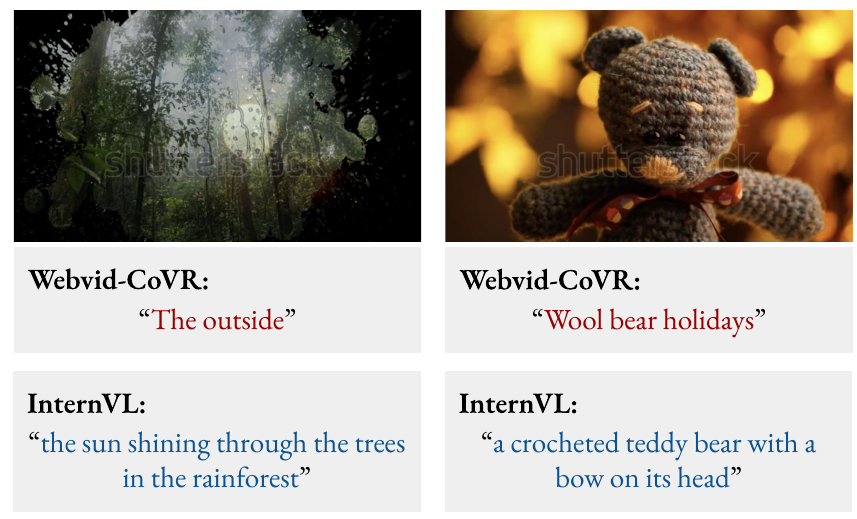}
\caption{Visual query samples along with their original captions provided by Ventura~\etal~\cite{Ventura-AAAI-2024} and captions generated by InternVL~\cite{Chen-CVPR-2024}. It is noticeable that captions generated by InternVL~\cite{Chen-CVPR-2024} are more closely aligned with the visual inputs compared to the original ones.}

\label{fig:caption_diff}
\end{figure}

\subsection{Ablation Results} 

\begin{figure*}[t]
  \includegraphics[width=0.999\linewidth]{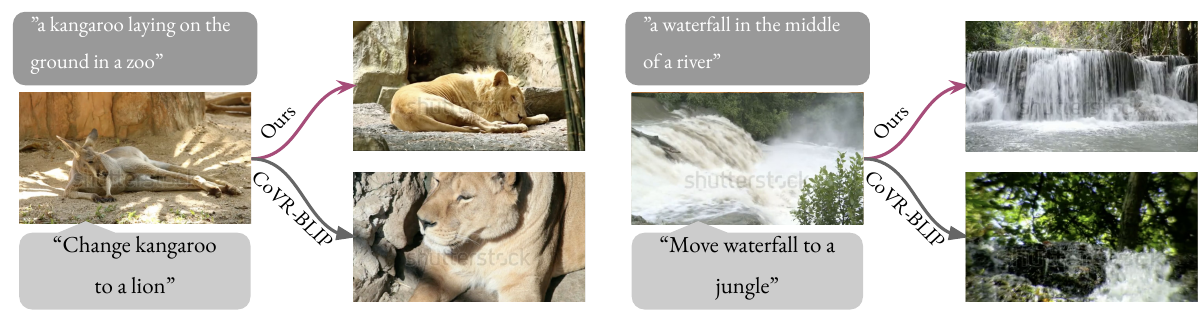}
\caption{Qualitative results comparing the results obtained using our framework and CoVR-BLIP~\cite{Ventura-AAAI-2024}. Our framework is able to retrieve the correct target given the multimodal input. The samples are extracted from WebVid-CoVR-Test~\cite{Ventura-AAAI-2024} and we display only the middle frame of the videos for clarity.}
\label{fig:qualitative}
\end{figure*}

\noindent
\textbf{Caption Contribution.} The use of the query visual caption leads to observable performance changes, as detailed in Table~\ref{tab:ablation_adding_caption}. While the results exhibit some variation depending on the presence of the caption, its inclusion consistently enhances the generalization ability of the embedding space.  
This is primarily because the caption serves as a stable modality, while the visual modality is changed from video (training) to images (inference), the text modality does not undergo a domain shift. This stability makes the cross-domain transfer significantly easier.

Moreover, adopting the BLIP-2 backbone further boosts the R@1 score on the CoVR task by nearly 1\% (from 62.99 to 63.93), indicating that the input caption also contributes positively to in-domain retrieval performance.

\begin{table} 
  \caption{Ablation results  when replacing the original \textit{input} captions (WebVid-CoVR) with captions generated using InternVL~\cite{Chen-CVPR-2024}. We report Composed Video Retrieval results on WebVid-CoVR-Test~\cite{Ventura-AAAI-2024} and the underlying VLM is BLIP~\cite{Li-ICML-2022}. The captions generated by InternVL~\cite{Chen-CVPR-2024} significantly outperform the original ones.}
  \label{tab:diff_caption}
  \centering
  \begin{tabular}{lccc}
    \hline
      \bf Caption & \bf R@1 & \bf R@5 & \bf R@10 \\
      \hline
       WebVid-CoVR~\cite{Ventura-AAAI-2024} & 52.62 & 77.11 & 84.98 \\
       InternVL~\cite{Chen-CVPR-2024} & \rc 63.50  & \rc 85.95 & \rc 91.59 \\
    \hline
  \end{tabular}
\end{table}

\noindent
\textbf{Input Caption Source.} To further investigate the effect of the \textit{input caption source}, we report results in Table~\ref{tab:diff_caption}, comparing the original captions provided by Ventura~\etal~\cite{Ventura-AAAI-2024} with those generated by InternVL~\cite{Chen-CVPR-2024}. The results reveal a notable performance gap between the two settings. This difference can be attributed to the fact that the original captions often fail to accurately describe the visual content, as illustrated in Figure~\ref{fig:caption_diff}. 

For example, where the original annotation offers a vague label like \textit{``The outside''} (Figure~\ref{fig:caption_diff} -- first image), InternVL generates a detailed scene description \textit{``the sun shining through the trees in the rainforest''}.
This increased granularity offered by the captions generated by InternVL allows our model to better align the visual and textual modalities, leading to superior retrieval accuracy

\subsection{Qualitative Results}

Qualitative comparisons with CoVR-BLIP~\cite{Ventura-AAAI-2024} are provided in Figure~\ref{fig:qualitative}. Our framework consistently retrieves target samples that accurately reflect the intended transformations (e.g., “Change kangaroo to a lion” or “Move waterfall to a jungle”). However, CoVR-BLIP sometimes struggles to integrate visual and textual cues. For instance, in the second example, it selects a scene with generic water flow, missing the core concept of a waterfall explicitly mentioned in the query. This indicates a limitation in grounding the textual instruction to the correct visual entity, despite the presence of related but semantically distinct content. Our method meanwhile accurately localizes the target concept ofa waterfall relocated to a jungle, demonstrating a better grasp of both visual grounding and textual intent.
 

%% file: sec/05_conclusions.tex
\section{Conclusions}
In this work, we propose a simple yet effective framework, \textbf{\ourmethod}, that bridges Composed Video Retrieval and Zero-Shot Composed Image Retrieval, demonstrating that even lightweight architectures can substantially enhance multimodal content understanding and retrieval when guided by well-informed design choices. By jointly leveraging additional caption representations, employing high-capacity pre-trained vision-language backbones, and updating the query text encoder during fine-tuning, our framework achieves a state-of-the-art R@1 score of 63.93\% on WebVid-CoVR-Test~\cite{Ventura-AAAI-2024}, while also generalizing effectively to cross-domain zero-shot CIR benchmarks such as FashionIQ~\cite{Wu-CVPR-2021} and CIRCO~\cite{Baldrati-ICCV-2023}.

A primary limitation of our current framework is its dependency on an external captioning model. However, given the rapid advancements and strong performance of modern Video LLMs, this reliance does not significantly bottleneck performance. 

%% file: main.bib
@String(CVPR= {IEEE Conf. Comput. Vis. Pattern Recog.})

@String(ICCV= {Int. Conf. Comput. Vis.})

@String(ECCV= {Eur. Conf. Comput. Vis.})

@String(ICLR = {Int. Conf. Learn. Represent.})

@String(AAAI = {AAAI})

@String(CVPR  = {CVPR})

@String(ICCV  = {ICCV})

@String(ECCV  = {ECCV})

@String(ICLR  = {ICLR})

@INPROCEEDINGS{Baldrati-ICCV-2023,
  author={Baldrati, Alberto and Agnolucci, Lorenzo and Bertini, Marco and Del Bimbo, Alberto},
  booktitle={2023 IEEE/CVF International Conference on Computer Vision (ICCV)}, 
  title={Zero-Shot Composed Image Retrieval with Textual Inversion}, 
  year={2023},
  volume={},
  number={},
  pages={15292-15301},  }

@article{Saito-CVPR-2023,
  title={Pic2Word: Mapping Pictures to Words for Zero-shot Composed Image Retrieval},
  author={Saito, Kuniaki and Sohn, Kihyuk and Zhang, Xiang and Li, Chun-Liang and Lee, Chen-Yu and Saenko, Kate and Pfister, Tomas},
  journal={CVPR},
  year={2023}
}

@inproceedings{Baldrati-CVPR-2023,
  title={Zero-Shot Composed Image Retrieval with Textual Inversion},
  author={Baldrati, Alberto and Agnolucci, Lorenzo and Bertini, Marco and Del Bimbo, Alberto},
  booktitle={Proceedings of the IEEE/CVF International Conference on Computer Vision},
  pages={15338--15347},
  year={2023}
}

@InProceedings{Radford-ICML-2021,
  title = 	 {Learning Transferable Visual Models From Natural Language Supervision},
  author =       {Radford, Alec and Kim, Jong Wook and Hallacy, Chris and Ramesh, Aditya and Goh, Gabriel and Agarwal, Sandhini and Sastry, Girish and Askell, Amanda and Mishkin, Pamela and Clark, Jack and Krueger, Gretchen and Sutskever, Ilya},
  booktitle = 	 {Proceedings of the 38th International Conference on Machine Learning},
  pages = 	 {8748--8763},
  year = 	 {2021}, 
  volume = 	 {139},
  series = 	 {Proceedings of Machine Learning Research},
  month = 	 {18--24 Jul},
  publisher =    {PMLR}, 
}

@article{Gu-TMLR-2024,
    title={CompoDiff: Versatile Composed Image Retrieval With Latent Diffusion},
    author={Geonmo Gu and Sanghyuk Chun and Wonjae Kim and HeeJae Jun and Yoohoon Kang and Sangdoo Yun},
    journal={Transactions on Machine Learning Research},
    issn={2835-8856},
    year={2024},
    url={https://openreview.net/forum?id=mKtlzW0bWc},
    note={Expert Certification}
}

@inproceedings{Chen-CVPR-2024,
  title={Internvl: Scaling up vision foundation models and aligning for generic visual-linguistic tasks},
  author={Chen, Zhe and Wu, Jiannan and Wang, Wenhai and Su, Weijie and Chen, Guo and Xing, Sen and Zhong, Muyan and Zhang, Qinglong and Zhu, Xizhou and Lu, Lewei and others},
  booktitle={Proceedings of the IEEE/CVF Conference on Computer Vision and Pattern Recognition},
  pages={24185--24198},
  year={2024}
}

@InProceedings{Lin-ECCV-2014,
author="Lin, Tsung-Yi
and Maire, Michael
and Belongie, Serge
and Hays, James
and Perona, Pietro
and Ramanan, Deva
and Doll{\'a}r, Piotr
and Zitnick, C. Lawrence", 
title="Microsoft COCO: Common Objects in Context",
booktitle="ECCV 2014",
year="2014", 
pages="740--755", 
}

@InProceedings{Li-CVPR-2025,
      title={Imagine and Seek: Improving Composed Image Retrieval with an Imagined Proxy}, 
      author={You Li and Fan Ma and Yi Yang},
      year={2025},
    booktitle = {Proceedings of the IEEE/CVF Conference on Computer Vision and Pattern Recognition (CVPR)},
    month     = {June}
}

@inproceedings{Gu-CVPR-2024,
    title={Language-only Training of Zero-shot Composed Image Retrieval},
    author={Gu, Geonmo and Chun, Sanghyuk and Kim, Wonjae and and Kang, Yoohoon and Yun, Sangdoo},
    year={2024},
    booktitle={Conference on Computer Vision and Pattern Recognition (CVPR)},
}

@article{Wu-CVPR-2021,
  title={The Fashion IQ Dataset: Retrieving Images by Combining Side Information and Relative Natural Language Feedback},
  author={Wu, Hui and Gao, Yupeng and Guo, Xiaoxiao and Al-Halah, Ziad  and Rennie, Steven and Grauman, Kristen and Feris, Rogerio},
  journal={CVPR},
  year={2021}
}

@inproceedings{Li-ICML-2022,
  title={BLIP: Bootstrapping Language-Image Pre-training for Unified Vision-Language Understanding and Generation},
  author={Junnan Li and Dongxu Li and Caiming Xiong and Steven C. H. Hoi},
  booktitle={International Conference on Machine Learning},
  year={2022}, 
}

@inproceedings{
    Karthik-ICLR-2024,
    title={Vision-by-Language for Training-Free Compositional Image Retrieval},
    author={Shyamgopal Karthik and Karsten Roth and Massimiliano Mancini and Zeynep Akata},
    booktitle={The Twelfth International Conference on Learning Representations},
    year={2024},
    url={https://openreview.net/forum?id=EDPxCjXzSb}
}

@article{Hummel-ECCV-2024,
  title={EgoCVR: An Egocentric Benchmark for Fine-Grained Composed Video Retrieval},
  author={Thomas Hummel and Shyamgopal Karthik and Mariana-Iuliana Georgescu and Zeynep Akata},
  journal={European Conference on Computer Vision (ECCV)},
  year={2024}
}

@article{Ventura-TPAMI-2024,
  title   = {{CoVR-2}: Automatic Data Construction for Composed Video Retrieval},
  author  = {Lucas Ventura and Antoine Yang and Cordelia Schmid and G{\"u}l Varol},
  journal = {IEEE TPAMI},
  year    = {2024}
}

@InProceedings{Thawakar-CVPR-2024,
    author    = {Thawakar, Omkar and Naseer, Muzammal and Anwer, Rao Muhammad and Khan, Salman and Felsberg, Michael and Shah, Mubarak and Khan, Fahad Shahbaz},
    title     = {Composed Video Retrieval via Enriched Context and Discriminative Embeddings},
    booktitle = {Proceedings of the IEEE/CVF Conference on Computer Vision and Pattern Recognition (CVPR)},
    month     = {June},
    year      = {2024},
    pages     = {26896--26906}
}

@inproceedings{Ventura-AAAI-2024,
  title     = {{CoVR}: Learning Composed Video Retrieval from Web Video Captions},
  author    = {Lucas Ventura and Antoine Yang and Cordelia Schmid and G{\"u}l Varol},
  booktitle = {AAAI},
  year      = {2024}
}

@article{Guo-CVPR-2021,
  title={Fashion IQ: A New Dataset Towards Retrieving Images by Natural Language Feedback},
  author={Xiaoxiao Guo and Hui Wu and Yupeng Gao and Steven J. Rennie and Rog{\'e}rio Schmidt Feris},
  journal={2021 IEEE/CVF Conference on Computer Vision and Pattern Recognition (CVPR)},
  year={2021},
  pages={11302--11312}
}

@inproceedings{Li-ICML-2023,
author = {Li, Junnan and Li, Dongxu and Savarese, Silvio and Hoi, Steven},
title = {BLIP-2: bootstrapping language-image pre-training with frozen image encoders and large language models},
year = {2023}, 
booktitle = {Proceedings of the 40th International Conference on Machine Learning},
articleno = {814},
numpages = {13}, 
}

@article{Wu-ICLR-2025, 
    title = {Learning Fine-Grained Representations through Textual Token Disentanglement in Composed Video Retrieval},
    author = {Yue, Wu and Zhaobo, Qi and Yiling, Wu and Junshu, Sun and Yaowei, Wang and Shuhui, Wang},
    journal = {ICLR},
    year = {2025}
}

@inproceedings{Vo-CVPR-2019,
  title={Composing Text and Image for Image Retrieval-An Empirical Odyssey},
  author={Vo, Nam and Jiang, Lu and Sun, Chen and Murphy, Kevin and Li, Li-Jia and Fei-Fei, Li and Hays, James},
  booktitle={CVPR},
  year={2019}
}

@inproceedings{Baldrati-CVPR-2022,
  title={Conditioned and Composed Image Retrieval Combining and Partially Fine-Tuning CLIP-Based Features},
  author={Baldrati, Alberto and Bertini, Marco and Uricchio, Tiberio and Del Bimbo, Alberto},
  booktitle={Proceedings of the IEEE/CVF Conference on Computer Vision and Pattern Recognition},
  pages={4959--4968},
  year={2022}
}

@inproceedings{Zhang-PMLR-2024, 
  title = 	 {{M}agic{L}ens: Self-Supervised Image Retrieval with Open-Ended Instructions},
  author =       {Zhang, Kai and Luan, Yi and Hu, Hexiang and Lee, Kenton and Qiao, Siyuan and Chen, Wenhu and Su, Yu and Chang, Ming-Wei},
  booktitle = 	 {Proceedings of the 41st International Conference on Machine Learning},
  pages = 	 {59403--59420},
  year = 	 {2024}, 
  volume = 	 {235},
  series = 	 {Proceedings of Machine Learning Research},
  month = 	 {21--27 Jul},
  publisher =    {PMLR}, 
}

@inproceedings{Yang-ICM-2024,
  title={Semantic Editing Increment Benefits Zero-Shot Composed Image Retrieval},
  author={Yang, Zhenyu and Qian, Shengsheng and Xue, Dizhan and Wu, Jiahong and Yang, Fan and Dong, Weiming and Xu, Changsheng},
  booktitle={Proceedings of the 32nd ACM International Conference on Multimedia},
  pages={1245--1254},
  year={2024}
}

@article{Zhou-Arxiv-2024,
  title={MegaPairs: Massive Data Synthesis For Universal Multimodal Retrieval},
  author={Zhou, Junjie and Liu, Zheng and Liu, Ze and Xiao, Shitao and Wang, Yueze and Zhao, Bo and Zhang, Chen Jason and Lian, Defu and Xiong, Yongping},
  journal={arXiv preprint arXiv:2412.14475},
  year={2024}
}

@article{Tang-arxiv-2024,
  title={Reason-before-Retrieve: One-Stage Reflective Chain-of-Thoughts for Training-Free Zero-Shot Composed Image Retrieval},
  author={Tang, Yuanmin and Qin, Xiaoting and Zhang, Jue and Yu, Jing and Gou, Gaopeng and Xiong, Gang and Ling, Qingwei and Rajmohan, Saravan and Zhang, Dongmei and Wu, Qi},
  journal={arXiv preprint arXiv:2412.11077},
  year={2024}
}

@inproceedings{Luo-WC-2025,
  title={ImageScope: Unifying Language-Guided Image Retrieval via Large Multimodal Model Collective Reasoning},
  author={Luo, Pengfei and Zhou, Jingbo and Xu, Tong and Xia, Yuan and Xu, Linli and Chen, Enhong},
  booktitle={The Web Conference 2025},
  year={2025}
}

@article{Levy-AAAI-2024,
 title={Data Roaming and Quality Assessment for Composed Image Retrieval},
  volume={38},
  number={4},
  journal={Proceedings of the AAAI Conference on Artificial Intelligence},
  author={Levy, Matan and Ben-Ari, Rami and Darshan, Nir and Lischinski, Dani},
  year={2024},
  month={Mar.},
  pages={2991-2999}
}

@article{rdk+23,
  title = {Filtering, Distillation, and Hard Negatives for Vision-Language Pre-Training},
  author = {Radenovic, Filip and Dubey, Abhimanyu and Kadian, Abhishek and Mihaylov, Todor and Vandenhende, Simon and Patel, Yash and Wen, Yi and Ramanathan, Vignesh and Mahajan, Dhruv},
  journal = {arXiv:2301.02280},
  year = {2023}
}

@inproceedings{thawakar2025beyond,
  title={Beyond simple edits: Composed video retrieval with dense modifications},
  author={Thawakar, Omkar and Demidov, Dmitry and Thawkar, Ritesh and Anwer, Rao Muhammad and Shah, Mubarak and Khan, Fahad Shahbaz and Khan, Salman},
  booktitle={Proceedings of the IEEE/CVF International Conference on Computer Vision},
  pages={20435--20444},
  year={2025}
}
